\pdfoutput=1
\documentclass[conference]{IEEEtran}
\IEEEoverridecommandlockouts

\usepackage{array}
\usepackage{multirow}
\usepackage{color}
\usepackage{algorithm}
\usepackage{tikz}
\usepackage{makecell}
\usetikzlibrary{shadows,arrows,shapes,positioning}
\usepackage{graphicx}
\usepackage[nointegrals]{wasysym} 

\newcolumntype{P}[1]{>{\centering\arraybackslash}p{#1}}
\newcolumntype{M}[1]{>{\centering\arraybackslash}m{#1}}
\usepackage{pifont}
\usepackage{enumitem}
\setlist{nolistsep}
\usepackage{algpseudocode}

\usepackage{subfig}

\usepackage[utf8]{inputenc} 
\usepackage[T1]{fontenc}    
\usepackage{booktabs}       
\usepackage{amsfonts}       
\usepackage{nicefrac}       
\usepackage{microtype}      
\usepackage{graphicx}
\usepackage{amssymb}
\usepackage{algorithm}
\usepackage{algpseudocode}
\usepackage{tikz}

\usepackage{stmaryrd}
\usepackage{trimclip}

\makeatletter
\DeclareRobustCommand{\shortto}{%
  \mathrel{\mathpalette\short@to\relax}%
}

\newcommand{\short@to}[2]{%
  \mkern2mu
  \clipbox{{.5\width} 0 0 0}{$\m@th#1\vphantom{+}{\shortrightarrow}$}%
  }
\makeatother

\usepackage{bbm}

\usepackage{cite}
\usepackage{amsmath,amssymb,amsfonts}
\usepackage{graphicx}
\usepackage{textcomp}
\usepackage{xcolor}
\def\BibTeX{{\rm B\kern-.05em{\sc i\kern-.025em b}\kern-.08em
    T\kern-.1667em\lower.7ex\hbox{E}\kern-.125emX}}

\begin{document}

\title{General Domain Adaptation Through Proportional Progressive Pseudo Labeling
\thanks{This research was supported in part by VMware and the NSF as part of SDI-CSCS award number 1700527.\\
978-1-7281-6251-5/20/\$31.00 ©2020 IEEE}
}

\author{\IEEEauthorblockN{Mohammad J. Hashemi}
\IEEEauthorblockA{\textit{Department of Computer Science} \\
\textit{University of Colorado Boulder}\\
Boulder, CO, USA \\
mohammad.hashemi@colorado.edu}
\and
\IEEEauthorblockN{Eric Keller}
\IEEEauthorblockA{\textit{Department of Electrical Computer and Energy Engineering} \\
\textit{University of Colorado Boulder}\\
Boulder, CO, USA \\
eric.keller@colorado.edu}
}

\maketitle

\begin{abstract}
Domain adaptation helps transfer the knowledge gained from a labeled source domain to an unlabeled target domain. During the past few years, different domain adaptation techniques have been published. One common flaw of these approaches is that while they might work well on one input type, such as images, their performance drops when applied to others, such as text or time-series. In this paper, we introduce Proportional Progressive Pseudo Labeling (PPPL), a simple, yet effective technique that can be implemented in a few lines of code to build a more general domain adaptation technique that can be applied on several different input types. At the beginning of the training phase, PPPL progressively reduces target domain classification error, by training the model directly with pseudo-labeled target domain samples, while excluding samples with more likely wrong pseudo-labels from the training set and also postponing training on such samples. Experiments on 6 different datasets that include tasks such as anomaly detection, text sentiment analysis and image classification demonstrate that PPPL can beat other baselines and generalize better.
\end{abstract}


\section{Introduction}
Deep neural networks have been used to approach many machine learning tasks such as object recognition, sentiment analysis, and anomaly detection, and they have achieved the state of the art results on those tasks and even surpassed humans capabilities \cite{delv, drive, twitter_sentiment, repo, DAGMM, kitsune, deep_ids_survey}. But, usually, in order to train these models, one needs to collect a very large labeled dataset. The problem is that labeling a huge dataset is expensive, time-consuming and needs a human in the loop. Possible workarounds to this issue include collecting a dataset similar to the target dataset that is already labeled because they had been used for an older task, or collecting a dataset that can easily be labeled, such as a synthetic dataset. Unfortunately, when we train a model on this labeled dataset, it doesn't work well on the target dataset if they are not coming from the same distribution and when there is a domain gap between them \cite{transferable_features, datashift_book}.
In some cases, when we have a very large and ongoing stream of inputs to the system, like when we want to detect network attacks based on the stream of packets that traverse through a network, this approach becomes more cumbersome as the input data distribution is constantly changing and also new attacks are introduced that may not exist in the training data which makes it even harder to detect them.
Ideally, for such a case, we would like to label a small portion of network traffic that also includes some attacks and be able to detect most types of attacks, including zero-day attacks, in the future without a need to directly label them.

Unsupervised domain adaptation methods are used to address such problems where there is a domain shift between data distributions of a labeled source domain and an unlabeled target domain. 
Recently, many different domain adaptation methods have been proposed \cite{can,cdan,SE,TAT,DA_ganin2016,DA_CyCADA2018,DA_JAN2017, dsne, Kurmi_2019_CVPR, MADA}. 
Despite different types of techniques used in recent approaches, one common flaw among them is that they don't generalize well among different types of inputs. Some of the methods such as \cite{DA_CyCADA2018,SE} are intrinsically designed for images, as they do image-to-image translation at the pixel level, or they need specific data augmentation that should be applied to them at the pixel level. Therefore, there is no straightforward way to apply these techniques to other input types such as text or time-series data. For other approaches, they either leverage adversarial loss \cite{cdan, DA_ganin2015, DA_ganin2016} or other techniques such as clustering \cite{can}. One common problem is that there is no guarantee preventing the wrong alignment of the target samples. In other words, target domain representations from one class can get aligned with another class during the domain adaptation, leading to lower performance of the model.  In addition, the complexity and the large number of hyper-parameters that some of these methods have weighs on this problem as there is no straightforward way to find the hyper-parameters that minimize the target error due to the lack of a labeled validation set for the target domain. Therefore, more complexity leads to less generalization.

In this paper, we introduce Proportional Progressive Pseudo Labeling (PPPL), a more general domain adaptation technique that works across different input types. PPPL assigns pseudo-labels to the target samples and trains the model directly with them. Key to PPPL is that it tries to minimize the number of target samples that will align with a wrong class by excluding uncertain samples from the training set at the beginning of the training procedure and progressively bring them back into the training loop with a weight proportional to their certainty. 
Further, we assume that we can guess the proportions of target samples that belong to each class. Note that while we don't know the individual labels in the target domain, the class proportions can be guessed in many cases. For example during an object recognition task based on the images collected from a camera deployed on an autonomous vehicle, replacing that camera with a new one can cause a domain shift, but will retain the class proportions. 
While other domain adaptation methods don't consider this condition, we show that enforcing class proportions during the training further decreases incorrect alignment of target samples that results in further improvement of overall system performance. 
Experiments on 6 different datasets (CIC-IDS2017 \cite{cic_dataset}, Yahoo \cite{yahoo}, Multi-Domain Sentiment \cite{MDS}, CIFAR-STL \cite{cifar,stl}, Office-31 \cite{office31} and Office-Home \cite{officehome}) including tasks such as anomaly detection, text sentiment analysis, and object recognition, demonstrates that our approach is superior to other baselines and generalizes better than them across different input types. Our experiments show that while PPPL is capable of improving the accuracy of image classifiers on visual domain adaptation tasks as good as state-of-the-art methods, it significantly outperforms them on other tasks with up to 62\% improvement for anomaly detection in network traffic based on the F1 score. Finally, with an ablation study, we demonstrate the necessity of each component in our method and with a sensitivity analysis, we show how robust our method is to the accuracy of class proportions that are guessed.
\section{Related Work}
Recent methods that are proposed for domain adaptation leverage a wide range of techniques to make a classifier get better results on the target domain. Some of them, such as CyCADA \cite{DA_CyCADA2018} and self-ensembling (SE) \cite{SE}, are designed for a specific type of input, namely images. In contrast, our work on PPPL can easily be applied to several different input types.

The second group of methods are those that don't require being applied on a specific input type as their focus is to mitigate the gap between the source and the target domain in the feature space at some intermediate layer of a deep network. The adversarial domain adaptation is the basic idea behind a large portion of recent approaches \cite{cdan, DA_ganin2015, DA_ganin2016}, in which the classifier is trained jointly with a domain discriminator like a GAN \cite{GAN}. The discriminator is trained to distinguish between source and target samples based on their representation captured from an intermediate layer of a deep network while the classifier itself is trained in a way to fool the discriminator that results in generating domain invariant features. Ganin et al. \cite{DA_ganin2016} proposed domain adversarial neural network (DANN) in which they augment a classifier with a domain discriminator and train them in an adversarial fashion by back-propagating the reverse gradients of the domain classifier to learn domain invariant representations. Long et al. \cite{cdan} designed conditional adversarial domain adaptation (CDAN) in which they condition the domain discriminator on the cross-covariance of domain-specific feature representations and classifier predictions, as well as on the uncertainty of the classifier to prioritize the discriminator on the easy to transfer samples. While we also model target domain uncertainty into our PPPL method, our approach differs from approaches like CDAN as they model the uncertainties into the domain discriminator.  Doing so makes those approaches deal with the complications of training GANs, whereas in our approach there is no discriminator and we model uncertainty directly into the classification loss.  This results in less complexity and greater generalization. 

Beyond adversarial domain adaptation methods, there are also other approaches like Contrastive Adaptation Network (CAN). Kang et al. \cite{can} mitigate the gap between the source and target domains at some feature space through the help of an alternating optimization method in which they initially cluster target samples into multiple different groups and then they assign some pseudo labels to them.  They then train the model by minimizing intra-class discrepancy and maximizing inter-class discrepancy.  Similar to CAN, we also assign some pseudo labels to the target samples, but unlike CAN, we don't use any clustering method.  We instead assign the pseudo labels directly based on the model predictions.  This, again, leads to less complexity and greater generalization.

\section{Design Insights}

Our goal is to train a deep network with the help of a labeled source domain in order to maximize the correct predictions on an unlabeled target domain. Formally, given a set of labeled samples known as source domain $S = \{(x^s_1,y^s_1), (x^s_2,y^s_2), ..., (x^s_{N_s},y^s_{N_s})\}$ such that $y_i \in Y=\{0,1,...,M-1\}$ and another set of unlabeled samples known as target domain $T = \{x^t_1, x^t_2, ..., x^t_{N_t} \}$ which come from two different data distributions our goal is to train a model $F: x \mapsto y$ to predict $\hat{y}^t_i \in Y$ to minimize the prediction error on the target domain. That is to say we want to minimize $\Sigma^{N_t}_{i=1} \mathbbm{1}(\hat{y}^t_i \neq y^t_i) $ where $y^{t}_i$ are the ground truth labels of the target domain samples.
In the rest of this section, we first provide some insights that helped us to design our approach. Then, in the following section, we explain how we incorporate these insights to design PPPL.

\begin{figure*}[t]
\centering
\subfloat[]{
\includegraphics[width=.32\linewidth]{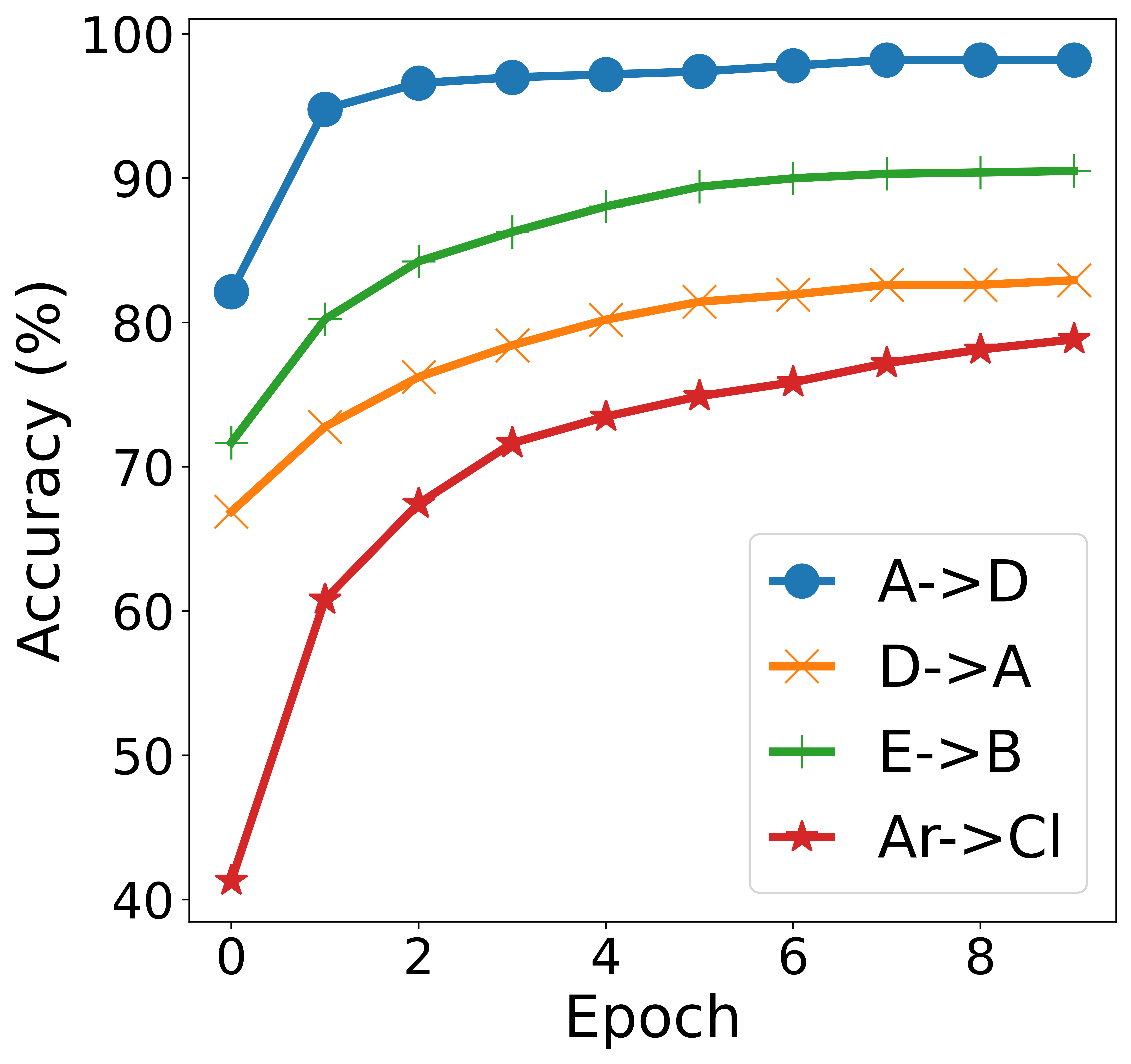}
}
\subfloat[]{
\includegraphics[width=.32\linewidth]{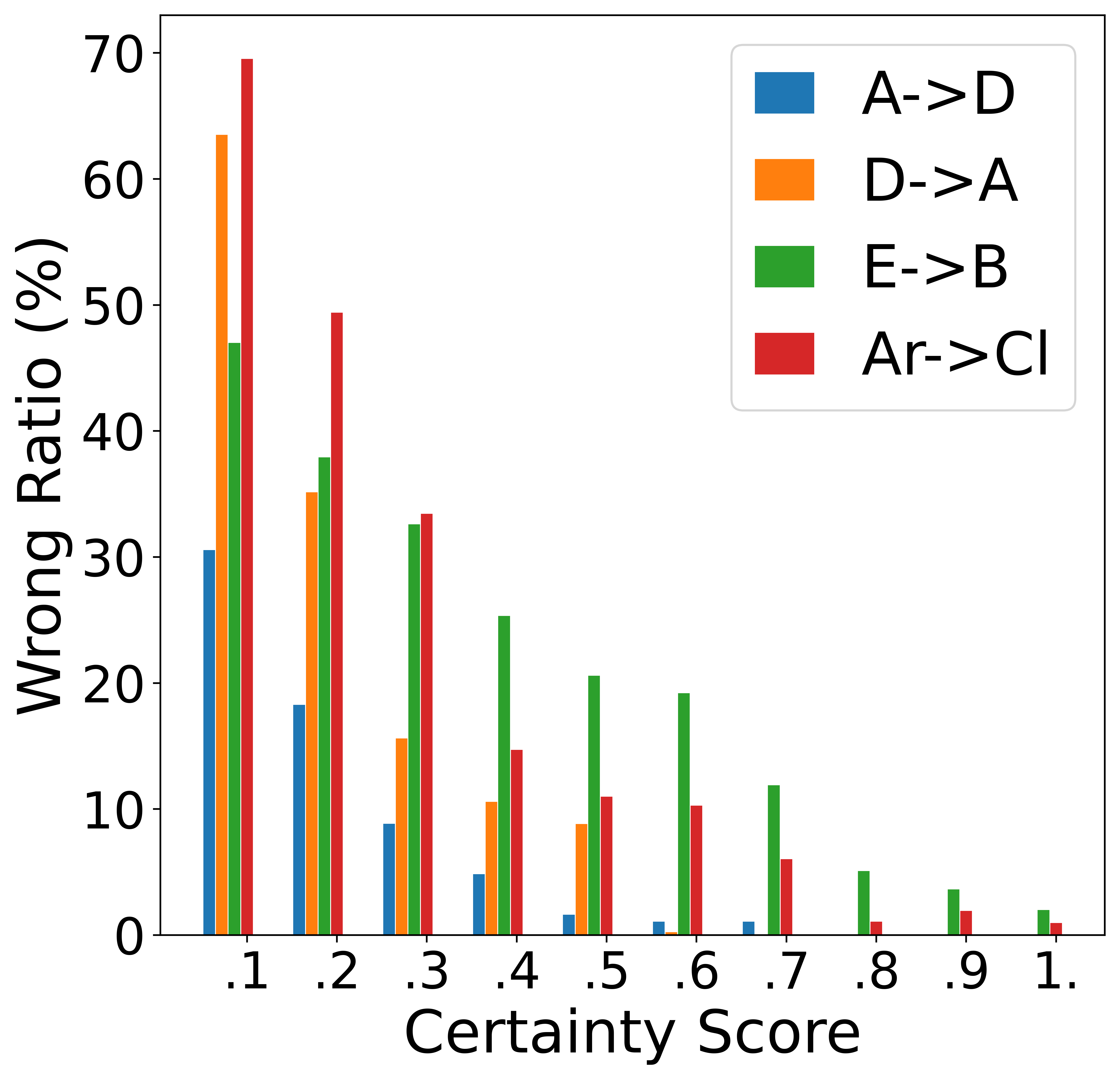}
}
\subfloat[]{
\includegraphics[width=.32\linewidth]{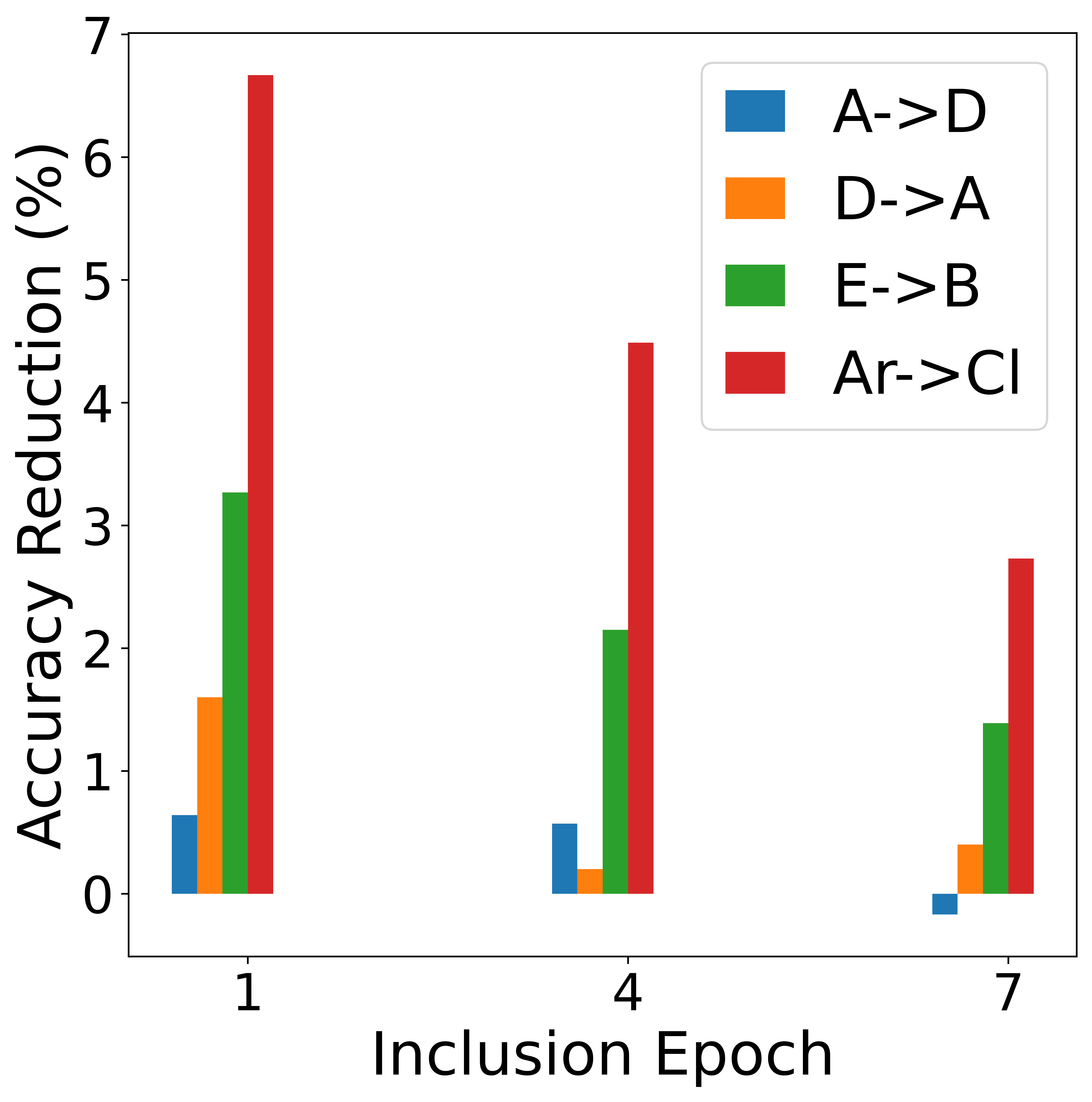}
}

\caption{a) Model accuracy on the target domain when trained only on correct predictions. b) The ratio of wrong predictions at different levels of model certainty. c) The negative impact of early training on wrong pseudo-labels. }
\label{fig:whatever}
\end{figure*}

\subsection{Insight 1 - when training with pseudo-labels mean square error is a better choice}
As we mentioned earlier, we want to assign pseudo labels directly based on model predictions to the target domain samples and train the model with them. For training, there are two choices here: We can feed the outputs of the final layer of the model to a Softmax function and train the model with the cross-entropy (CE) loss. We can also directly minimize the distance of the final layer outputs and the one-hot encoding of the labels with mean square error loss (MSE). We argue that for such a setting, MSE is a better choice. First, consider the Softmax function which is defined as follows:\par
{\centering $ \sigma(z)_i = \frac{e^{z_i}}{\Sigma_{j=0}^{M-1}e^{z_j}}$ for $i = 0,1,...,M-1$\par}
\noindent where M is the number of classes. Note that because of the nature of this function there is no one-to-one mapping between the probabilities (outputs of the Softmax layer) and the logits (inputs of the Softmax layer). That is to say, many different points in the logit space can be mapped to a single vector of probabilities. For example, when there is only 2 classes, both of the points $[1,100]$ and $[200,299]$ will be mapped to the $[\frac{1}{1+e^{99}},\frac{e^{99}}{1+e^{99}}]$. This means that potentially, the points that belong to the same class can form multiple different clusters in the logit space. More specifically, the target domain samples and the source domain samples that share the same label or pseudo label can fall into different clusters. On the other hand, when we train the model with MSE, the points that have real or pseudo label $C_i$ will fall into one single cluster very close to the point $[0,...,0,1,0,...0]$ where the i-th index is 1 after we train the model on them. This characteristic is more desirable as it mitigates the domain gap between the source and the target samples at logit space and forces the model to learn features in the earlier layers of the network that leads to indistinguishable representations at the logit space between the source and the target domain samples. From another point of view, if we would have a domain discriminator to distinguish between the source and the target samples logits it could be completely fooled. Therefore by using MSE loss in this setting we get the same advantages of adversarial domain adaptation techniques without being worried about the complications of training GANs.

\subsection{Insight 2 - training only on correct samples gradually reduces the target error}
Consider a model pretrained on the source domain. Also, consider some samples (e.g. A, B and C) from the target domain that belongs to the same class (e.g. class 1) and fall into the close proximity of each other at some middle representation of the network. Assume that we assign pseudo-labels to these samples directly based on model predictions. Suppose that some of these pseudo-labels are correct and some are wrong (e.g. $\hat{y}_A=1, \hat{y}_B=2, \hat{y}_C=2 $). If we exclude wrong samples (B, C) and train the model only on correct samples (A), then the model learns that the points (B, C) in close proximity of these points (A) are also more likely to be from the same class (class 1) and potentially some of them will get a correct pseudo-label in the next iteration. This effect gets propagated to the points in close proximity of B and C in the next iteration. Therefore, excluding wrong samples and training only on correct samples gradually reduces the target error.

To further show this, in Figure \ref{fig:whatever}(a) we demonstrate the results of such training on 4 different domain adaptation tasks namely Amazon $\rightarrow$ DSLR and DSLR $\rightarrow$ Amazon from the Office-31 dataset, Art $\rightarrow$ Clipart from the OfficeHome dataset, and Electronics $\rightarrow$ Books from the Multi-Domain Sentiment (MDS) dataset. As can be seen in all cases, the model progressively learns to predict target domain samples more accurately. These results are better than the results of any other domain adaptation approach by a large margin and it also generalizes well across different input types. In other words, we can significantly reduce the target error only by assigning pseudo labels to the target samples directly based on the model predictions and training the classifier with them. We just somehow need to determine in which cases the model is wrong to exclude them from the training procedure. 

\subsection{Insight 3 - an uncertainty metric can guide which predictions are wrong}
Unfortunately, there is no straightforward way to know which of the model predictions are correct and which ones are wrong on the target domain as we don't know the target domain's labels.  But, among all samples that are predicted as the same class, there is a relation between the model's certainty and the chance of wrong predictions.  We capture the model's certainty with the difference between the two largest scores that the model outputs for each sample and call it the certainty score. This difference becomes smaller as in some intermediate representation of the inputs the points get further away from the same-labeled source points, falling into sub-spaces that are not well explored by the model or when they fall in close proximity to other points with different labels meaning getting closer to the decision boundaries. Thus, in such cases, it becomes more likely to get predicted wrongly.  As can be seen in Figure \ref{fig:whatever}(b), in general when certainty score decreases among samples that are given the same pseudo-label, a larger portion of predictions becomes wrong. For this figure, we first trained the model on the source domain for each of the aforementioned tasks with MSE loss. Then for each class $C$ the ratio of wrong predictions to all of the target samples that are predicted as $C$ and their certainty scores fall into the interval $[\frac{i-1}{10}, \frac{i}{10}]$ is calculated. For each task, the i-th bar demonstrates the average of such ratio across all the classes.

\subsection{Insight 4 - the timing of inclusion of a wrong-prediction matters}
While we can predict better, we cannot know for sure which predictions are correct, and therefore it might be inevitable we assign some wrong pseudo-labels to some of the target samples and train the model on them. But one thing that is important is the time when we train the model on the target samples with the wrong pseudo-labels. We argue that the early inclusion of such samples into the training procedure deteriorates the model's performance on the target domain more than later inclusion. This is because of the same phenomenon that we discussed in insight 2: When we train a model on a target sample with a wrong pseudo-label, it would be more likely for the model to assign that wrong label to the points that are in close proximity of that wrong sample.  Then, this wrong label propagates to the neighborhood of these newly affected samples in the next iteration. Therefore, the earlier we train the model on a wrong sample, the further its impact will propagate, the more model's accuracy deteriorates on the target domain. This can be seen in Figure \ref{fig:whatever}(c). For this figure, for each of the tasks mentioned in the second insight, we trained the model the same way we discussed but also we included some samples with wrong pseudo-labels into the training procedure at different epochs (epochs 1,4,7 and 10). The size of the wrong pseudo-labeled samples in each task is equal to 10\% of the target domain sample size. For all of the cases, we trained the model for 10 epochs. In this figure, we illustrate the change in the model accuracy when the wrong pseudo-labeled samples were included in the training loop at epoch 1, 4, and 7 in comparison with when they were included at epoch 10. Therefore a larger bar shows a greater decrease in the model's accuracy. As can be seen, the earlier the model got trained on the wrong samples the more its accuracy is decreased. For example, for the Art $\rightarrow$ Clipart task if we include the wrong pseudo-labeled samples at the first epoch of training the final accuracy will be almost 7\% lower than when we postpone their inclusion to epoch 10.

\section{Proportional Progressive Pseudo Labeling (PPPL)}
Based on the insights we discussed we designed our approach. In a nutshell, based on the second insight we know that if we use a model pre-trained on the source domain and assign pseudo-labels to the target samples with it and exclude the wrong pseudo-labels the model progressively gets better. Also, based on the first insight we know that for such a setting using MSE loss is better than CE loss. Unfortunately, since we don't know target labels we can't find out exactly for which cases the pseudo-labels are wrong but based on the third insight we know that the ratio of wrong predictions increases as the model certainty decreases. In addition, based on the fourth insight we know that it is better to postpone training the model on such samples. 

Algorithm \ref{alg1} describes our approach in more detail. The inputs are $F$ which is the model pretrained on the source domain, $X_s$ which is the set of all source domain samples, $Y_s$ which is the set of all the source domain labels, $X_t$ which is the set of all target domain samples and $CP_t$ which is the set of target class proportions that are guessed or known from other sources.
We first train the model ($F$) on the source domain samples $(X_s,Y_s)$ with the MSE loss function (based on insight 1).
Then in each iteration of the algorithm, we first get the score which is a vector with size M (M is the number of available classes) for each of the target samples (line 4) and assign a pseudo-label to that sample based on its largest score.  (line 5). Then for all of the target samples, we calculate the "certainty score" (line 6) and then assign a weight value to each of the target samples based on its certainty score (line 7).  This weight will be used later during training to control the impact of each sample on the model parameters. 

\begin{algorithm} 
\caption{Proportional Progressive Pseudo-Labeling}\label{alg1}
\begin{algorithmic}[1]
\Procedure{PPPL}{$F,X_s,Y_s,X_t,CP_t$} 
    \For {$i \gets 1$ to $45$}
    \State $N \gets 10+ 2\times i$
    \State $S_t \gets F(X_t)$
    \State $PL_t \gets argmax(S_t)$
    \State $CS_t \gets CalcCertaintyScore(S_t)$
    \State $W_t \gets CalculateWeight(CS_t,PL_t,N)$
    \State $X^\prime_t,Y^\prime_t,W^\prime_t \gets Exclude(X_t,PL_t,W_t,CP_t)$
    \State $X^\prime_s,Y^\prime_s,W_s \gets Select(X_s,Y_s)$
    \State $Train(F,X^\prime_s,Y^\prime_s,W_s,X^\prime_t,Y^\prime_t,W^\prime_t)$
    \EndFor

\EndProcedure
\end{algorithmic}
\end{algorithm}

The function $CalculateWeight(CS_t,PL_t,N)$, first groups all of the samples that are assigned the same label. Then, for each group, it assigns a weight between $[0.2 - 1.0]$ to $N\%$ of the samples and $0$ to the rest of the samples of that group. Within each group, the weights are monotonically assigned based on the certainty scores such that a sample with a larger certainty score will be assigned a larger weight. More specifically, if the number of samples that fall into top $N\%$ for a given group is $L_c$ then the weights for those samples are calculated as follows: $w_j = \frac{1}{t_j}$ where $t_j = 1 + \frac{4}{L_c}\times j$ in which $j \in [0,L_c)$ and $w_j$ is assigned to the j-th sample with the largest certainty score. 

\begin{figure}
  \centering
  \includegraphics[width=\linewidth]{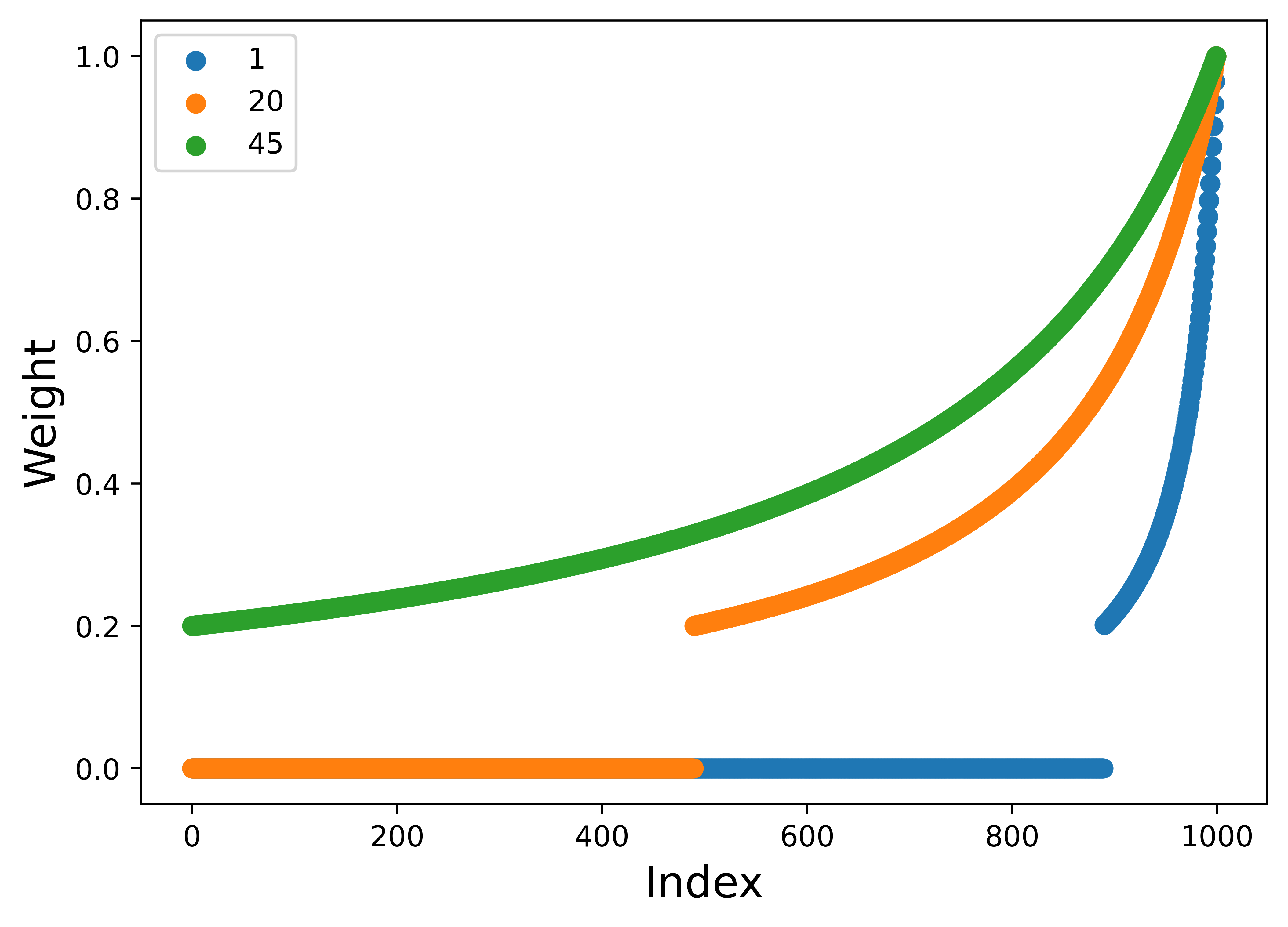}
  \caption{The illustration of weights assigned to the target samples with the same pseudo-label.}
  \label{fig:weights}
\end{figure}

For better illustration, in Figure \ref{fig:weights} we show the weights that will be assigned to the target samples with the same pseudo-label at epoch 1, 20 and 45 of our method. We assumed that the size of this group would remain in 1000 during the whole training. As can be seen, for the first epoch only 10\% of the samples will be assigned with a weight of more than 0, which essentially means that the other 90\% samples that have a lower certainty score will be excluded from training in the first epoch. At epoch 20, 50\% of the samples will be included in the training and at the final epoch, all the samples will be included. 
Using this weighting strategy decreases the chance of training on wrong samples on the early epochs by excluding the less certain samples (designed based on insights 3 and 4). Also for the samples that will be included at each epoch, we assign a smaller weight to the less certain samples to decrease the impact of those that are assigned wrong pseudo-labels and potentially are among included samples on the model parameters (designed based on insight 3). 

There is also another way we try to exclude wrong samples which is based on guessed class proportions (line 8). In the $Exclude$ function, for each label, we first calculate the ratio of samples with that pseudo-label to the whole target domain sample size. Then we exclude some samples for classes that their ratio is higher than its corresponding guessed ratio because it means that the model predicted samples with that label more than what we expect. Therefore, we expect that some of these predictions to be wrong. So, from each class, we keep excluding the most uncertain predictions until the ratio of remaining samples becomes equal to our expected ratio (designed based on insights 2 and 3). 

In addition to the target samples, we also select some samples randomly from the source domain at each epoch to train the model on them. We do this because in the initial phases of our algorithm we only include a small portion of the target samples and we don't want to make the model over-fit on them. We also assign a weight equal to 1 to these source samples as all of their labels are correct. Finally with the combination of pseudo-labeled target samples and these source samples we train the model with MSE loss (designed based on insight 1). More specifically, the loss function we use for training is as follows: $\frac{1}{N}\sum^N_{i=1} w_i||F(x_i) - y_i||_2^2$ in which $y_i$ is the one-hot encoding of (pseudo)label assigned to sample $x_i$ and $N$ is the total number of included samples at the current epoch.

\section{Evaluation}
In this section, we evaluate PPPL for doing domain adaptation on 6 different datasets. For two of them (CIC-IDS2017 and Yahoo) we want to do anomaly detection based on time-series, in one of them (MDS) we want to do sentiment analysis based on text input, and for the other three (Office-31, Office-Home, CIFAR-STL) we want to do object recognition in image inputs. 

\begin{table*}
  \caption{Results of all methods on the CICIDS-2017 dataset. The numbers reported in this table are F1 scores.}
  \label{cic_table}
  \centering
  \begin{tabular}{cccccccc}
    \toprule
    Method     & Tu	$\rightarrow$ We     & Tu	$\rightarrow$ Th & We	$\rightarrow$ Tu & We	$\rightarrow$ Th & Th	$\rightarrow$ Tu & Th	$\rightarrow$ We & Avg \\
    \midrule
    Only-Src & 0.055 & 0.215 & 0.028 & 0.087 & 0.010 & 0.016 & 0.068\\
    CDAN & 0.007 & 0.004 & 0.006 & 0.013 & 0.000 & 0.025 & 0.009\\
    CAN & 0.662 & 0.169 & 0.123 & 0.333 & 0.000 & 0.005 & 0.215\\
    \midrule
    PPPL & \textbf{0.973} & \textbf{0.708} & \textbf{0.855} & \textbf{0.712} & \textbf{0.784} & \textbf{0.964} & \textbf{0.833} \\
    \bottomrule
  \end{tabular}
\end{table*}

\begin{table*}
  \caption{Results of all methods on the CIFAR-STL and Yahoo datasets. We reported accuracies for the CIFAR-STL dataset and F1 scores for the Yahoo dataset.}
  \label{cifar_yahoo_table}
  \centering
  \begin{tabular}{cccc|cccc}
    \toprule
    Method     & CIFAR $\shortrightarrow$  STL & STL $\shortrightarrow$  CIFAR & Avg & Real $\shortrightarrow$  Syn. & Syn. $\shortrightarrow$  Real & Avg \\
    \midrule
    Only-Src & 76.0 & 61.3 & 68.6 & 0.441 & 0.081 & 0.261\\
    CDAN & 77.6 & 63.3 & 70.5 & 0.448 & 0.049 & 0.249\\
    CAN & 76.6 & 55.5 & 66.0 & 0.016 & 0.187 & 0.102 \\
    \midrule
    PPPL & \textbf{79.6} & \textbf{69.7} & \textbf{74.6} & \textbf{0.624} & \textbf{0.285} & \textbf{0.454}\\
    \bottomrule
  \end{tabular}
\end{table*}

\subsection{Datasets}

\textbf{CIC-IDS2017}
dataset consists of network traces of different network attacks. The whole dataset which is collected over a week contains millions of packets and each packet is labeled as either \textit{benign} or \textit{malicious}. The task in this dataset is to detect malicious packets. The type of attacks that are carried out on different days differs from each other. The Tuesday attacks are FTP-Patator and SSH-Patator. The Wednesday attacks are DoS Slowloris, DoS Slowhttptest, DoS Hulk, DoS GoldenEye and Heartbleed. The Thursday attacks are Web attacks and Infiltration.
We consider the traffic collected in each of these days as one domain and define 6 domain adaptation tasks between each pair of them  (e.g. Tu$\rightarrow$We, Tu$\rightarrow$Th). In order to classify the packets in this dataset, we first preprocessed them with the same method described in \cite{repo}. Given this preprocessing method, the Tu, We and Th domains contain 573,544, 685,241 and 462,031 samples, respectively. Also, all of these domains are imbalanced. The percentage of malicious traffic in the Tu, We, and Th domains are 2.2\%, 18.2\% and 2.7\%, respectively.

\textbf{Yahoo} dataset contains real and synthetic time-series and the goal is to detect anomalous points in these time-series. The real domain consists of metrics of different Yahoo services and reflects the status of the Yahoo system and the synthetic dataset is generated artificially. Both of the domains are extremely imbalanced. There are 67 time-series in the real domain and 100 ones in the synthetic domain and each of them contains multiple anomalous points. We preprocessed this dataset as follows: For each of the time-series, we first calculated the difference between each point and its adjacent point and then normalized all the points in that time-series to have a mean of 0 and standard deviation of 1. Then, we concatenated each point with the 511 points that came before it and fed this vector to the classifier as the input. Given this preprocessing method, the real domain contains 60,629 samples where 1,563 of them are anomalous and the artificial domain contains 91,000 samples where 466 of them are anomalous.

\textbf{Multi-Domain Sentiment (MDS)} is widely used to evaluate domain adaptation methods built for sentiment-analysis based on text inputs. It contains 27,677 product reviews from amazon.com about four product domains: books (B), DVDs (D), electronics (E) and kitchen appliances (K) and therefore 12 different tasks. The goal is to classify the reviews into positive and negative classes. For each domain, 2,000 reviews are named labeled and around 4,000 are named unlabeled. For each task, after Bag of Words (BoW) preprocessing, during the training of each method, we used 2,000 labeled reviews from the source domain and both of the labeled and unlabeled reviews from the target domain (without their labels). 

\textbf{CIFAR-STL} dataset is a combination of images from 9 overlapping classes of CIFAR-10 and STL. There are 45,000 images in the CIFAR-10 domain and 4,500 images in the STL domain.

\textbf{Office-31}
is widely used for the evaluation of visual domain adaptation methods and contains 4,652 images in 3 different domains known as Amazon (A), DSLR (D) and Webcam (W) and 31 different classes. There are 6 domain adaptation tasks between each pair of domains (e.g. A$\rightarrow$D, D$\rightarrow$W, ...) and we evaluate PPPL and other baselines on all of the 6 tasks.

\textbf{Office-Home} is another dataset used for visual domain adaptation that is harder than Office-31. This dataset contains 15,500 images in 65 classes and there are in 4 different domains known as Art (Ar), Clipart (Cl), Product (Pr) and Real-World (Rw). Therefore for this dataset, 12 different tasks exist between each pair of domains.

\begin{table*}
  \caption{Accuracy (\%) for all the 12 tasks of the MDS dataset. }
  \label{mds_table}
  \centering
    \begin{tabular}{>{\centering\arraybackslash}m{1.5cm}m{0.5cm}m{0.5cm}m{0.5cm}m{0.5cm}m{0.5cm}m{0.5cm}m{0.5cm}m{0.5cm}m{0.5cm}m{0.5cm}m{0.5cm}m{0.5cm}m{0.5cm}}
    \toprule
    Method     &B$\shortrightarrow$ D&B$\shortrightarrow$ E& B$\shortrightarrow$ K & D$\shortrightarrow$ B & D$\shortrightarrow$ E & D$\shortrightarrow$ K & E$\shortrightarrow$ B & E$\shortrightarrow$ D & E$\shortrightarrow$ K & K$\shortrightarrow$ B & K$\shortrightarrow$ D & K$\shortrightarrow$ E & Avg \\
    \midrule
    Only-Src & 81.0 & 73.0 & 75.6 & 78.0 & 74.2 & 78.5 & 71.4 & 72.3 & 85.7 & 73.2 & 74.0 & 86.3 & 76.9 \\
    CDAN & \textbf{83.3} & 80.5 & 82.4 & 79.0 & 68.3 & 82.1 & 61.8 & 65.9 & 86.0 & 65.5 & 65.4 & 84.3 & 
 75.4\\
    CAN & 79.3 & 77.8 & 80.5 & 75.6 & 76.2 & 81.4 & 73.9 & 74.4 & 85.3 & 73.9 & 70.9 & 83.1 & 77.7 \\
    \midrule
    PPPL & 83.2 & \textbf{84.2} & \textbf{86.0} & \textbf{81.9} & \textbf{84.4} & \textbf{87.1} & \textbf{75.5} & \textbf{80.3} & \textbf{89.9} & \textbf{77.1} & \textbf{79.6} & \textbf{89.2} & \textbf{83.2} \\
    \bottomrule
  \end{tabular}
\end{table*}

\begin{table*}
  \caption{Accuracy (\%) for all the six tasks of Office-31 based on ResNet-50. }
  \label{office31_table}
  \centering
  \begin{tabular}{cccccccc}
    \toprule
    Method     & A	$\rightarrow$ D     & A	$\rightarrow$ W & D	$\rightarrow$ A & D	$\rightarrow$ W & W	$\rightarrow$ A & W	$\rightarrow$ D & Avg \\
    \midrule
    Only-Src & 82.1 & 79.4 & 66.9 & 98.1 & 66.3 & 99.8 & 82.1\\
    CDAN & 92.9 & 94.1 & 71.0 & 98.6 & 69.3 & 100.0 & 87.7\\
    CAN & 95.0 & 94.5 & \textbf{78.0} & 99.1 & 77.0 & 99.8 & 90.6 \\
    \midrule
    PPPL & \textbf{95.0} & \textbf{96.1} & 77.8 & \textbf{99.2} & \textbf{77.3} & \textbf{100.0} & \textbf{90.9} \\
    \bottomrule
  \end{tabular}
\end{table*}

\begin{table*}
  \caption{Accuracy (\%) for all the 12 tasks of Office-Home based on ResNet-50. }
  \label{officehome_table}
  \centering

    \begin{tabular}{>{\centering\arraybackslash}m{1.5cm}m{0.6cm}m{0.6cm}m{0.6cm}m{0.6cm}m{0.6cm}m{0.6cm}m{0.6cm}m{0.6cm}m{0.6cm}m{0.6cm}m{0.6cm}m{0.6cm}m{0.6cm}}

    \toprule
    Method     &Ar$\shortrightarrow$ Cl&Ar$\shortrightarrow$ Pr& Ar$\shortrightarrow$ Rw & Cl$\shortrightarrow$ Ar & Cl$\shortrightarrow$ Pr & Cl$\shortrightarrow$ Rw & Pr$\shortrightarrow$ Ar & Pr$\shortrightarrow$ Cl & Pr$\shortrightarrow$ Rw & Rw$\shortrightarrow$ Ar & Rw$\shortrightarrow$ Cl & Rw$\shortrightarrow$ Pr & Avg \\
    \midrule
    Only-Src & 41.3 & 67.8 & 75.0 & 56.9 & 65.2 & 67.2 & 53.7 & 39.0 & 74.5 & 66.2 & 43.2 & 78.1 & 60.7 \\
    CDAN & 50.7 & 70.6 & 76.0 & 57.6 & 70.0 & 70.0 & 57.4 & 50.9 & 77.3 & 70.9 & 56.7 & 81.6 & 65.8\\
    CAN & 60.0 & 79.0 & 81.3 & 68.2 & 78.9 & 78.3 & 67.7 & 57.1 & 83.1 & 74.3 & \textbf{62.9} & 84.9 & 73.0 \\
    \midrule
    PPPL & \textbf{62.9} & \textbf{80.2} & \textbf{82.2} & \textbf{70.0} & \textbf{80.8} & \textbf{80.7} & \textbf{69.5} & \textbf{58.4} & \textbf{83.6} & \textbf{77.2} & 62.0 & \textbf{85.8} & \textbf{74.4} \\
    \bottomrule
  \end{tabular}
\end{table*}

\subsection{Experiments}
\textbf{Baselines:} We compare PPPL with two other baselines which are among the best domain adaptation techniques to the best of our knowledge: CAN \cite{can} and CDAN \cite{cdan}. CAN is state-of-the-art for the Office-31 dataset. Kang et al. in their evaluation showed that CAN outperforms many other baselines such as Domain Adversarial Neural Network (DANN)  \cite{DA_ganin2015,DA_ganin2016}, Joint Adaptation Network (JAN) \cite{DA_JAN2017}, Multi-adversarial Domain Adaptation (MADA) \cite{MADA}, Deep Adaptation Network (DAN) \cite{DAN} and Self Ensembling (SE) \cite{SE}. Also, CDAN is among the best adversarial domain adaptation techniques that we are aware of. Long et al. in their evaluation showed that CDAN outperforms many other methods such as DAN \cite{DAN},  Residual Transfer Networks (RTN) \cite{rtn}, DANN \cite{DA_ganin2015, DA_ganin2016}, Adversarial Discriminative Domain Adaptation (ADDA) \cite{adda}, JAN \cite{DA_JAN2017} and CyCADA \cite{DA_CyCADA2018}. Therefore by comparing our approach with CAN and CDAN and outperforming them, the superiority of PPPL over many other baselines can be inferred.

\textbf{Setup:} 
For evaluation of PPPL on the CIC-IDS2017, Yahoo, MDS and CIFAR-STL datasets we trained a model from scratch. After preprocessing, we trained a two-layer fully-connected network $(2048\rightarrow 2)$ for CIC-IDS2017, a four-layer fully-connected network $(2048\rightarrow 2048 \rightarrow 256 \rightarrow 2)$ for Yahoo and another four-layer fully-connected network $(2048\rightarrow 2048 \rightarrow 2048 \rightarrow 2)$ for the MDS dataset. For the CIFAR-STL dataset, we trained the same convolutional network used in \cite{SE} for this dataset. For the Office-31 and Office-Home datasets we used ResNet-50 \cite{resnet} pretrained on ImageNet \cite{imagenet} as the feature extractor. We removed the final layer of Resnet-50 model and added 4 new layers at the end $(4096\rightarrow 4096 \rightarrow 4096 \rightarrow \#classes)$. In all of the datasets, we first trained the model on the labeled source samples and then we did domain adaptation with labeled source samples and unlabeled target samples according to the Algorithm \ref{alg1}. \footnote{For more details, please see https://github.com/s-mohammad-hashemi/pppl.} Also, in all of the cases here we assumed that we know the class proportions in the target domain accurately. In the sensitivity analysis section, we show how inaccurate class proportions impact the results.

\textbf{Results:} CAN and CDAN perform very poorly on the anomaly detection tasks. The results on the CIC-IDS2017 dataset are reported in Table \ref{cic_table} which are the F1 scores calculated for each case. As can be seen, on average PPPL is 61.8\% better than the best baseline we compared with. Table \ref{cifar_yahoo_table} shows the results of all methods on CIFAR-STL and Yahoo datasets. For the Yahoo dataset, because we have a very imbalanced dataset, we also reported F1 scores. Note that on average, our approach notably improves model performance on both datasets compared to training only on the source domain samples (6\% improvement for the CIFAR-STL dataset and 19.3\% improvement for the Yahoo dataset). Whereas, on average CAN performs even worse than just training on the source domain on both datasets. CDAN also performs worse than training on the source domain for the Yahoo dataset and just slightly better than training a model using only source domain samples on the CIFAR-STL dataset. The same problem can be observed for the MDS dataset. The results for this dataset are reported in Table \ref{mds_table}. Note that CDAN performs worse than training only on the source domain samples and CAN just slightly works better than it, while PPPL outperforms it by 6.3\%. 
In addition to outperforming other baselines in tasks such as anomaly detection and sentiment analysis, PPPL can still work as good as other baselines on the Office-31 and Office home datasets where our baselines demonstrate their best performance. The results on the Office-31 dataset are reported in Table \ref{office31_table}. On this dataset, PPPL works as good as CAN and even slightly better. The results on the Office-Home dataset are reported in Table \ref{officehome_table}. PPPL outperforms other baselines on this dataset. Our results are 1.4\% better than CAN and 8.6\% better than CDAN.
Therefore these results confirm that our approach generalizes better across different input types.

\subsection{Analysis}
\textbf{Ablation Study:} To show the necessity of each component of our method, we compare PPPL with the following alternative training strategies: A1) We replace MSE with CE loss. A2) We include all the target samples into the training loop from the first epoch. That is to say we modify line 7 of Algorithm \ref{alg1} to become $W_t \gets CalculateWeight(CS_t,PL_t,100)$. A3) Instead of class-aware weight assignment in $CalculateWeight$ function, we assign weights to target samples without grouping them based on their assigned pseudo-label and group them only based on their certainty scores. A4) We don't adjust the training set based on target class proportions. In other words, we remove line 8 from Algorithm \ref{alg1}. We make these comparisons for the following tasks: A $\rightarrow$ D from the Office-31 dataset, B $\rightarrow$ E from the MDS dataset and Syn $\rightarrow$ Real from the Yahoo dataset. Table \ref{ablation_table} shows the results. Note that on average, all of the alternative training strategies perform worse than PPPL.

\begin{table}

\caption{Alternative training strategies.}
  \label{ablation_table}
  \centering
    \begin{tabular}{>{\centering\arraybackslash}m{1.1cm}m{0.4cm}m{0.4cm}m{0.4cm}m{0.4cm}m{0.7cm}}
    \toprule

    Task  & A1  & A2  & A3  & A4  & PPPL  \\
    \midrule
    A $\rightarrow$ D  &  86.6  &  89.4  &  91.4  &  89.2  &  95.0  \\
    B $\rightarrow$ E  &  81.7  & 74.3  &  76.3  &  79.3  &  84.2  \\
    S $\rightarrow$ R &  0.187 &  0.259 &  0.095 &  0.282 &  0.285 \\
    \bottomrule
  \end{tabular}
  \end{table}

\begin{table}
  \caption{Training PPPL with enforcing different CP.}
  \label{sensitivity_table}
  \centering
    \begin{tabular}{>{\centering\arraybackslash}m{1.3cm}m{0.4cm}m{0.4cm}m{0.4cm}m{0.4cm}m{0.4cm}m{0.7cm}}

    \toprule

    Task & 10\%  & 20\%  & 30\%  & S. & T. & CP Diff.  \\ 
    \midrule
    A $\rightarrow$ D  &  96.6 &  95.0 &  94.8 &  91.4 & 95.0 & .32    \\
    A $\rightarrow$ W  &  95.5 &  95.5 &  95.8 &  95.0 & 96.1 & .21    \\
    D $\rightarrow$ A  &  77.7  &  77.0  &  74.3  &  74.7  & 77.8 & .32  \\
    Ar $\rightarrow$ Cl   &  61.5  &  61.6  &  62.3  &  57.9 & 62.9 & .47    \\ 
    S $\rightarrow$ R &  0.264 &  0.264 &  0.266 &  0.263 & 0.285 & .04  \\
    \bottomrule
  \end{tabular}
\end{table}

\textbf{Sensitivity Analysis:} During the evaluation of our method, we assumed that we know the exact values of class proportions of the target domain. Now we want to see how much error our method can tolerate in the class proportions. The bottom line is that we can always enforce source domain class proportions during training with PPPL as we can calculate them with the help of source labels. Therefore in Table \ref{sensitivity_table} we show how our method performs when we enforce source domain class proportions (column 5) and when there is 10\%, 20\% and 30\% error (column 2,3 and 4) in our guessed values compared to real target domain class proportions for five different tasks. In this table, for the anomaly detection task, the error is calculated for the anomalous class. That is to say for $E \in \{0.1,0.2,0.3\}$ anomalous class proportion is modified such that $|CP_a - \hat{CP}_a| = E\times CP_a$. For the other tasks, the error is calculated based on all the classes. More specifically, the guessed class proportions are chosen such that $\Sigma^M_{i=1} |CP_i - \hat{CP}_i| = E$ where $CP_i$ is the real class proportion for i-th class and $\hat{CP}_i$ is the guessed one. The last column of this table also shows the difference between the source domain and target domain class proportions which are calculated with this formula for all of the tasks. Note that for most of the tasks in the table, the algorithm's performance gets affected only slightly when there is up to 30\% error in the guessed class proportions. Also, note that just by enforcing source class proportions, we can still get good results when the difference between the source and target CPs is small (e.g. A $\rightarrow$ W task). Therefore our algorithm will be a good choice when we can estimate target class proportions with a small error or when the source and target class proportions are very similar.

\section{Conclusion}
In this paper, we proposed Proportional Progressive Pseudo Labeling (PPPL), a simple and novel method for domain adaptation that generalizes better than other methods across different input types. PPPL aims to progressively reduce the error on the target domain by assigning pseudo-labels to the target domain samples and training the model with them while excluding samples with more likely wrong pseudo-labels from the training set and also postponing training on such samples. Experiments on multiple different tasks confirm the superiority of our approach compared to other strong baselines.

\bibliographystyle{IEEEtran}
\bibliography{9_references}
\end{document}